\documentclass[10pt,final]{article}

\usepackage{ranlp}
\usepackage{epsf}
\usepackage{times}

\newif\ifpdf
\ifx\pdfoutput\undefined \else
  \ifx\pdfoutput\relax
  \else
    \ifcase\pdfoutput
    \else
      \pdftrue
    \fi
  \fi
\fi

\usepackage{t1enc}
\usepackage{graphicx}
\usepackage{tabularx}
\usepackage{url}
\usepackage{abbrevs}
\usepackage{color}

\let\em\itswitch

\providecommand{\ie}{\textit{i.e.} }%
\providecommand{\citeN}[1]{\cite{#1}}%

\newabbrev\tdt{Topic Detection and Tracking}
\newabbrev\mds{Multi-document Summarization}[MDS]
\newabbrev\rst{\emph{Rhetorical Structure Theory}}[RST]
\newabbrev\cst{Cross-do\-cu\-ment Structure Theory (CST)}[CST]
\newabbrev\nlg{Natural Language Generation (NLG)}[NLG]
\newabbrev\IE{Information Extraction (IE)}[IE]
\newabbrev\nerc{Named Entity Recognition and Classification (NERC)}[NERC]
\newabbrev\nes{Named Entities (NEs)}[NEs]
\newabbrev\muc{MUC}
\newabbrev\ml{Machine Learning}

\title{Summarizing Reports on Evolving Events;\\ Part I: Linear Evolution}
\author{Stergos D. Afantenos \and Vangelis Karkaletsis\\
        National Center for Scientific Research (NCSR) ``Demokritos'', Greece\\
        \texttt{\{stergos,vangelis\}@iit.demokritos.gr}
        \AND
        Panagiotis Stamatopoulos\\
        Department of Informatics, University of Athens, Greece\\
        \texttt{takis@di.uoa.gr}}

\begin{document}

\maketitle


\begin{abstract}
We present an approach for summarization from multiple documents which report
on events that evolve through time, taking into account the different document
sources. We distinguish the evolution of an event into \emph{linear} and
\emph{non-linear}. According to our approach, each document is represented by a
collection of messages which are then used in order to instantiate the
cross-document relations that determine the summary content. The paper presents
the summarization system that implements this approach through a case study on
linear evolution.
\end{abstract}

\ResetAbbrevs{All}

\section{Introduction}
With the advent of the Internet, access to many sources of information has now
become much more easier. One problem that arises though from this fact is that
of the information overflow. Imagine, for example, that someone wants to keep
track of an event that is being described on various news sources, over the
Internet, as it evolves through time. The problem is that there exist a
plethora of news sources that it becomes very difficult for someone to compare
the different versions of the story in each source. Furthermore, the Internet
has made it possible now to have a rapid report of the news, almost immediately
after they become available. Thus, in many situations it is extremely difficult
to follow the rate with which the news are being reported. In such cases, a
text summarizing the reports from various sources on the same event, would be
handy. In this paper we are concerned with the automatic creation of summaries
from multiple documents which describe an event that evolves through time. Such
a collection of documents usually contains news reports from various sources,
each of which provides novel information on the event as it evolves through
time. In many cases the sources will agree on the events that they report and
in some others they will adopt a different viewpoint presenting a slightly
different version of the events or possibly disagreeing with each other. Such a
collection of documents can, for example, be the result of a \tdt system
\cite{Allan&al98:TDTFinRep}.

The identification of similarities and differences between the documents is a
major aspect in \mds
\cite{Mani01,Afantenos&al.05:MedicalSurvey,Afantenos&al.05:NLUCS}.
\cite{Mani&Bloedorn99}, for example, identify similarities and differences
among \emph{pairs} of isolated documents by comparing the graphs that they
derive from each document, which are based heavily on various lexical criteria.
Our approach, in contrast, does not take into consideration isolated pairs of
documents, but instead tries to identify the similarities and differences that
exist between the documents, taking into account the time that the incidents
occurred and the document source. This enables us to distinguish the document
relations into \emph{synchronic} and \emph{diachronic} ones. In the synchronic
level we try to identify the similarities and differences that exist between
the various sources. In the diachronic level, on the other hand, we try to
identify similarities and differences across time focusing on each source
separately.

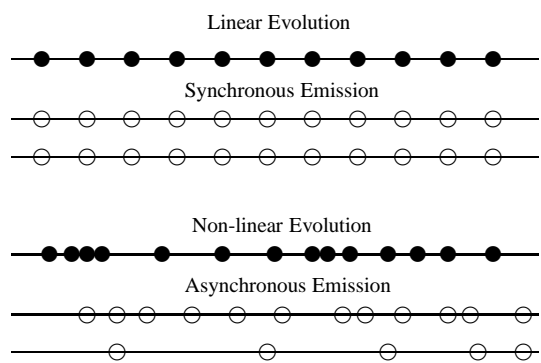
\begin{figure}[htb]
  \begin{center}
    \setlength{\unitlength}{4144sp}%
\begingroup\makeatletter\ifx\SetFigFont\undefined%
\gdef\SetFigFont#1#2#3#4#5{%
  \reset@font\fontsize{#1}{#2pt}%
  \fontfamily{#3}\fontseries{#4}\fontshape{#5}%
  \selectfont}%
\fi\endgroup%
\begin{picture}(3174,2078)(214,-1734)
{\color[rgb]{0,0,0}\thinlines
\put(406,-286){\circle{90}}
}%
{\color[rgb]{0,0,0}\put(676,-286){\circle{90}}
}%
{\color[rgb]{0,0,0}\put(946,-286){\circle{90}}
}%
{\color[rgb]{0,0,0}\put(1216,-286){\circle{90}}
}%
{\color[rgb]{0,0,0}\put(1756,-286){\circle{90}}
}%
{\color[rgb]{0,0,0}\put(2026,-286){\circle{90}}
}%
{\color[rgb]{0,0,0}\put(2296,-286){\circle{90}}
}%
{\color[rgb]{0,0,0}\put(2566,-286){\circle{90}}
}%
{\color[rgb]{0,0,0}\put(2836,-286){\circle{90}}
}%
{\color[rgb]{0,0,0}\put(3106,-286){\circle{90}}
}%
{\color[rgb]{0,0,0}\put(3106,-511){\circle{90}}
}%
{\color[rgb]{0,0,0}\put(2836,-511){\circle{90}}
}%
{\color[rgb]{0,0,0}\put(2566,-511){\circle{90}}
}%
{\color[rgb]{0,0,0}\put(2296,-511){\circle{90}}
}%
{\color[rgb]{0,0,0}\put(2026,-511){\circle{90}}
}%
{\color[rgb]{0,0,0}\put(1756,-511){\circle{90}}
}%
{\color[rgb]{0,0,0}\put(1486,-511){\circle{90}}
}%
{\color[rgb]{0,0,0}\put(1216,-511){\circle{90}}
}%
{\color[rgb]{0,0,0}\put(946,-511){\circle{90}}
}%
{\color[rgb]{0,0,0}\put(676,-511){\circle{90}}
}%
{\color[rgb]{0,0,0}\put(406,-511){\circle{90}}
}%
{\color[rgb]{0,0,0}\put(1486,-286){\circle{90}}
}%
{\color[rgb]{0,0,0}\put(2431,-286){\line( 1, 0){945}}
}%
{\color[rgb]{0,0,0}\put(2431,-511){\line( 1, 0){945}}
}%
{\color[rgb]{0,0,0}\put(226,-286){\line( 1, 0){2205}}
}%
{\color[rgb]{0,0,0}\put(226,-511){\line( 1, 0){2205}}
}%
{\color[rgb]{0,0,0}\put(451,-1096){\circle*{90}}
}%
{\color[rgb]{0,0,0}\put(586,-1096){\circle*{90}}
}%
{\color[rgb]{0,0,0}\put(676,-1096){\circle*{90}}
}%
{\color[rgb]{0,0,0}\put(766,-1096){\circle*{90}}
}%
{\color[rgb]{0,0,0}\put(1126,-1096){\circle*{90}}
}%
{\color[rgb]{0,0,0}\put(1801,-1096){\circle*{90}}
}%
{\color[rgb]{0,0,0}\put(2026,-1096){\circle*{90}}
}%
{\color[rgb]{0,0,0}\put(2116,-1096){\circle*{90}}
}%
{\color[rgb]{0,0,0}\put(2251,-1096){\circle*{90}}
}%
{\color[rgb]{0,0,0}\put(226,-1096){\line( 1, 0){2250}}
}%
{\color[rgb]{0,0,0}\put(2476,-1096){\line( 1, 0){900}}
}%
{\color[rgb]{0,0,0}\put(1486,-1096){\circle*{90}}
}%
{\color[rgb]{0,0,0}\put(2656,-1096){\circle*{90}}
}%
{\color[rgb]{0,0,0}\put(2836,-1096){\circle*{90}}
}%
{\color[rgb]{0,0,0}\put(3106,-1096){\circle*{90}}
}%
{\color[rgb]{0,0,0}\put(2476,-1096){\circle*{90}}
}%
{\color[rgb]{0,0,0}\put(856,-1681){\circle{90}}
}%
{\color[rgb]{0,0,0}\put(1756,-1681){\circle{90}}
}%
{\color[rgb]{0,0,0}\put(2476,-1681){\circle{90}}
}%
{\color[rgb]{0,0,0}\put(3016,-1681){\circle{90}}
}%
{\color[rgb]{0,0,0}\put(3286,-1681){\circle{90}}
}%
{\color[rgb]{0,0,0}\put(2476,-1681){\line( 1, 0){900}}
}%
{\color[rgb]{0,0,0}\put(226,-1681){\line( 1, 0){2250}}
}%
{\color[rgb]{0,0,0}\put(406, 74){\circle*{90}}
}%
{\color[rgb]{0,0,0}\put(676, 74){\circle*{90}}
}%
{\color[rgb]{0,0,0}\put(946, 74){\circle*{90}}
}%
{\color[rgb]{0,0,0}\put(1216, 74){\circle*{90}}
}%
{\color[rgb]{0,0,0}\put(1486, 74){\circle*{90}}
}%
{\color[rgb]{0,0,0}\put(1756, 74){\circle*{90}}
}%
{\color[rgb]{0,0,0}\put(2026, 74){\circle*{90}}
}%
{\color[rgb]{0,0,0}\put(2296, 74){\circle*{90}}
}%
{\color[rgb]{0,0,0}\put(2566, 74){\circle*{90}}
}%
{\color[rgb]{0,0,0}\put(2836, 74){\circle*{90}}
}%
{\color[rgb]{0,0,0}\put(3106, 74){\circle*{90}}
}%
{\color[rgb]{0,0,0}\put(676,-1456){\circle{90}}
}%
{\color[rgb]{0,0,0}\put(856,-1456){\circle{90}}
}%
{\color[rgb]{0,0,0}\put(1036,-1456){\circle{90}}
}%
{\color[rgb]{0,0,0}\put(1306,-1456){\circle{90}}
}%
{\color[rgb]{0,0,0}\put(1846,-1456){\circle{90}}
}%
{\color[rgb]{0,0,0}\put(1576,-1456){\circle{90}}
}%
{\color[rgb]{0,0,0}\put(2206,-1456){\circle{90}}
}%
{\color[rgb]{0,0,0}\put(2341,-1456){\circle{90}}
}%
{\color[rgb]{0,0,0}\put(2566,-1456){\circle{90}}
}%
{\color[rgb]{0,0,0}\put(2836,-1456){\circle{90}}
}%
{\color[rgb]{0,0,0}\put(2971,-1456){\circle{90}}
}%
{\color[rgb]{0,0,0}\put(3286,-1456){\circle{90}}
}%
{\color[rgb]{0,0,0}\put(2431, 74){\line( 1, 0){945}}
}%
{\color[rgb]{0,0,0}\put(2476,-1456){\line( 1, 0){900}}
}%
{\color[rgb]{0,0,0}\put(226, 74){\line( 1, 0){2205}}
}%
{\color[rgb]{0,0,0}\put(226,-1456){\line( 1, 0){2250}}
}%
\put(1396,254){\makebox(0,0)[lb]{\smash{{\SetFigFont{8}{9.6}{\familydefault}{\mddefault}{\updefault}{\color[rgb]{0,0,0}Linear Evolution}%
}}}}
\put(1306,-961){\makebox(0,0)[lb]{\smash{{\SetFigFont{8}{9.6}{\familydefault}{\mddefault}{\updefault}{\color[rgb]{0,0,0}Non-linear Evolution}%
}}}}
\put(1261,-1321){\makebox(0,0)[lb]{\smash{{\SetFigFont{8}{9.6}{\rmdefault}{\mddefault}{\updefault}{\color[rgb]{0,0,0}Asynchronous Emission}%
}}}}
\put(1261,-151){\makebox(0,0)[lb]{\smash{{\SetFigFont{8}{9.6}{\rmdefault}{\mddefault}{\updefault}{\color[rgb]{0,0,0}Synchronous Emission}%
}}}}
\end{picture}%
    \caption{Linear and Non-linear evolution}\label{fig:evolution}
  \end{center}
\end{figure}

Another twofold distinction that we made through our study
\cite{Afantenos&al.05:NLUCS} concerns the type of evolution of an event,
distinguishing between \emph{linear} and \emph{non-linear} evolution, and the
rate of emission of the various news sources, distinguishing between
\emph{synchronous} and \emph{asynchronous} emission of reports.
Figure~\ref{fig:evolution} depicts the major incidents for two different
events: a linearly evolving event with synchronous emission and a non-linearly
evolving one with asynchronous emission of reports. Whereas in the linearly
evolving events the main incidents happen in constant and possibly predictable
quanta of time,\footnote{This means that if the first news story $q_0$ comes at
moment $t_0$, then we can assume that for each source the story $q_n$ will come
at time $t_n = t_0 + n*t$, where $t$ is the constant amount of time that it
takes for the news to appear.} in the non-linear events we can make no
predictions as to when the next incident will occur. As you can see in
Figure~\ref{fig:evolution} we can have within a small amount of time an
explosion of incidents followed by a long time of sparse incidents, etc.

In order to represent the various incidents that are described in each
document, we introduce the notion of \emph{messages}. Messages are composed
from a name, which reflects the type of the incidents, and a list of arguments,
which take their values from the domain ontology. Additionally, they have
associated with them the \emph{time} that the message refers to, as well as the
document \emph{source}.

The distinction between linear and non-linear evolution affects mainly the
\emph{synchronic relations}, which are used in order to identify the
similarities and differences between two messages from different sources, at
about the same time. In the case of linear evolution all the sources report in
the same time. Thus, in most of the cases, the incidents described in each
document refer to the time that the document was published. Yet, in some cases
we might have temporal expressions in the text that modify the time that a
message refers to. In such cases, before establishing a synchronic relation, we
should associate this message with the appropriate time-tag. In the case of
non-linear evolution, each source reports at irregular intervals, possibly
mentioning incidents that happened long before the publication of the article,
and which another source might have already mentioned in an article published
earlier. In this case we shouldn't rely any more to the publication of an
article, but instead rely on the \emph{time} tag that the messages have (see
section~\ref{sec:definitions}). Once this has been performed, we should then
establish a \emph{time window} in which we should consider the messages, and
thus the relations, as synchronic.

In the following section, we make more concrete and formal the notion of the
messages and relations. In section~\ref{sec:case_study} we briefly present our
methodology and describe its implementation through a particular case study.
Section~\ref{sec:related} presents in more detail the related work, and
section~\ref{sec:future} concludes presenting ongoing work on \emph{non-linear}
summarization and our future plans.

\section{Some Definitions}\label{sec:definitions}
In our approach
\cite{Afantenos&al.05:NLUCS,Afantenos&al.04:SETN} the major building blocks for
representing the knowledge on a specific event are: the \emph{ontology} which
encodes the basic entity types (concepts) and their instances; the
\emph{messages} for representing the various incidents inside the document; and
the \emph{relations} that connect those messages across the documents. More
details are given below.

\paragraph{Ontology.} For the purposes of our work, a domain ontology should be
built. The ontology we use is a taxonomic one, incorporating \emph{is-a}
relations, which are later exploited by the messages and the relations.

\paragraph{Messages.} In order to capture what is represented by several
textual units, we introduce the notion of \emph{messages}. A message is
composed from four parts: its \emph{type}, a list of \emph{arguments} which
take their values from the concepts of the domain \emph{ontology}, the
\emph{time} that the message refers, and the \emph{source} of the document that
the message is contained. In other words, a message can be defined as follows:
\begin{center}
  \texttt{message\_type ( arg$_1$, $\ldots$ , arg$_n$ )}\\
  where \texttt{arg}$_i$ $\in$ Domain Ontology
\end{center}
Each message $m$ is accompanied by the time ($m$\texttt{.time}) that it refers
and its source ($m$\texttt{.source}). Concerning the source, this is inherited
by the source of the document that contains the message. Concerning the time of
the message, it is inherited by the publication time of the document, unless
there exists a temporal expression in the text that modifies the time that a
message refers. In this case, we should interpret the time-tag of the message,
in relation to that temporal expression. A message definition may also be
accompanied by a set of \emph{constraints} on the values that the arguments can
take. We would like also to note that messages are similar structures (although
simpler ones) with the templates used in the
\muc.\footnote{\url{http://www.itl.nist.gov/iaui/894.02/related_projects/muc/proceedings/muc_7_toc.html}}
An example of a message definition will be given in the case study we present
in section~\ref{sec:case_study}.

\paragraph{Relations.} In order to define a relation in a domain we have to
provide a \emph{name} for it, and describe the conditions under which it will
hold. The name of the relation is in fact \emph{pragmatic} information, which
we will be able to exploit later during the generation of the summary. The
conditions that a relation holds are simply some rules which describe the
\emph{temporal distance} that two messages should have (0 for synchronic and
more than 1 for diachronic) and the characteristics that the arguments of the
messages should exhibit in order for the relation to hold.

Furthermore, it is crucial to note here the importance that time and source
position have on the relations, apart from the values of the messages'
arguments. Suppose, for example, that we have two identical messages. If they
have the same temporal tag, but belong to different sources, then we have an
\textsl{agreement} relation. If, on the other hand, they come from the same
source but they have chronological distance one, then we speak of a
\textsl{stability} relation. Finally, if they come from different sources and
they have chronological distance more than two, then we have no relation at
all. We also do not have a relation if the messages have different sources and
different chronological distances. Thus we see that, apart from the
characteristics that the arguments of a message pair should exhibit, the source
and temporal distance also play a role for that pair to be characterized as a
relation. In section~\ref{sec:case_study} we will give concrete examples of
messages and relations for a particular case study.

\section{A Case Study of Linear Evolution}\label{sec:case_study}
The methodology was originally presented in
\cite{Afantenos&al.04:SETN}. It involves four stages:
\begin{enumerate}
  \item Corpus collection
  \item Creation of a domain ontology
  \item Specification of the messages
  \item Specification of the relations
\end{enumerate}

The topic we have chosen is that of the descriptions of football matches. In
this domain, we have several events that evolve; for example, the performance
of a player or a team as the championship progresses. According to the
definitions we have given, the evolution of this domain is \emph{linear}. The
reason for this is that we have a match each week which is then being described
by several sources.

As our methodology requires, in order to create multi-document summaries of
evolving events, we have to provide some knowledge of the domain to the system.
This knowledge is provided through the ontology and the specification of the
messages and the relations, following the four steps described above.

\subsection{Domain Knowledge}

\paragraph{Corpus Collection.}
We manually collected descri\-ptions of football matches, from various sources,
for the period 2002-2003 of the Greek football championship. The language used
in the documents was also Greek. This championship contained 30 rounds. We
focused on the matches of a certain team, which were described by three
sources. So, we had in total 90 documents.

\paragraph{Ontology Creation.}
An excerpt of the taxonomic ontology we have created is shown in
Figure~\ref{fig:ontology}.

\begin{figure}[thb]
\footnotesize
\begin{verbatim}
 Degree               Card
 Person                 Yellow
   Referee              Red
   Assistant Referee  Team
   Linesman           Temporal Concept
   Coach                Minute
   Player               Duration
   Spectators             First Half
     Viewers              Second Half
     Organized Fans       Delays
 Round                    Whole Match
\end{verbatim}\normalsize
\caption{An excerpt from the domain ontology}\label{fig:ontology}
\end{figure}

\paragraph{Messages' Specifications.}
We concentrated in the most important events, that is on events that evolve
through time, or events that a user would be interested in knowing. At the end
of this process we concluded on the following set of 23 message types:

\smallskip\noindent \footnotesize \textit{Absent, Behavior, Block, Card, Change,
Comeback, Conditions, Expectations, Final\_Score, Foul, Goal\_Cancelation,
Hope\_For, Injured, Opportunity\_Lost, Penalty, Performance, Refereeship,
Satisfaction, Scorer, Successive\_Victories, Superior, System\_Selection, Win}
\normalsize

\smallskip\noindent An example of full message specifications is shown in
Figure~\ref{fig:msgs}. We should note that this particular message type is not
accompanied by constraints. Also, associated with it we have the \texttt{time}
and \texttt{source} tags.

\begin{figure}[thb]
  \small\textbf{performance} (entity, in\_what, time\_span, value)\\
  \noindent
  \begin{tabularx}{2.3in}{Xll}
    entity     &:& player or team\\
    in\_what   &:& Action Area\\
    time\_span &:& Minute or Duration\\
    value      &:& Degree
  \end{tabularx}\normalsize
  \caption{An example of message specifications}\label{fig:msgs}
\end{figure}

\paragraph{Specification of the Relations.}
We identified twelve cross-document relations, six on the synchronic and six on
the diachronic axis (see Table~\ref{table:relations}).

\begin{table}[thb]
\centering
\begin{tabularx}{\columnwidth}{lX}
\textsl{\textbf{Diachronic Relations \hspace{0.2in}}} &
\textsl{\textbf{Synchronic Relations}}\\
\hline
-- \small\textsc{Positive Graduation} & -- \small\textsc{Agreement}\\
-- \small\textsc{Negative Graduation} & -- \small\textsc{Near Agreement}\\
-- \small\textsc{Stability}           & -- \small\textsc{Disagreement}\\
-- \small\textsc{Repetition}          & -- \small\textsc{Elaboration}\\
-- \small\textsc{Continuation}        & -- \small\textsc{Generalization}\\
-- \small\textsc{Generalization}      & -- \small\textsc{Preciseness}\\
\end{tabularx}
\caption{Synchronic and Diachronic Relations in the Football
Domain}\label{table:relations}
\end{table}

Since this was a pilot-study during which we examined mostly the feasibility of
our methodology, we limited the study of the cross-document relations, in those
ones that connect the \emph{same} message types. Thus both the synchronic and
the diachronic relations connect the same types, although further studies might
reveal that different message types can be connected with some sort of
relations. Furthermore, concerning the \emph{diachronic} relations we limited
our study in relations that have chronological distance only
one.\footnote{Chronological distance zero makes the relations synchronic.}
Examples of such specifications for the message type \texttt{performance} are
shown in Figure~\ref{fig:relSpecs}.

\begin{figure}[thb]
  \scriptsize\textbf{Performance}\\
  Assuming we have the following two messages:
  \begin{quote}
    performance$_1$ (entity$_1$, in\_what$_1$, time\_span$_1$, value$_1$)\\
    performance$_2$ (entity$_2$, in\_what$_2$, time\_span$_2$, value$_2$)
  \end{quote}
  Then we have a Diachronic relation if
  \begin{quote}
    (performance$_1$\texttt{.time} $<$ performance$_2$\texttt{.time}) and\\
    (performance$_1$\texttt{.source} = performance$_2$\texttt{.source})
  \end{quote}
  and a Synchronic relation if
  \begin{quote}
    (performance$_1$\texttt{.time} = performance$_2$\texttt{.time}) and\\
    (performance$_1$\texttt{.source} $\neq$ performance$_2$\texttt{.source})
  \end{quote}
  More specifically, we have the following Synchronic and Diachronic relations:
  \begin{quote}
    \textsl{Diachronic Relations}
      \begin{itemize}
        \item \textbf{Positive Graduation} iff\\
              (entity$_1$ = entity$_2$) and (in\_what$_1$ = in\_what$_2$) and
              (time\_span$_1$ = time\_span$_2$) and (value$_1$ $<$ value$_2$)
        \item \textbf{Stability} iff\\
              (entity$_1$ = entity$_2$) and (in\_what$_1$ = in\_what$_2$) and
              (time\_span$_1$ = time\_span$_2$) and (value$_1$ = value$_2$)
        \item \textbf{Negative Graduation} iff\\
              (entity$_1$ = entity$_2$) and (in\_what$_1$ = in\_what$_2$) and
              (time\_span$_1$ = time\_span$_2$) and (value$_1$ $>$ value$_2$)
      \end{itemize}
    \textsl{Synchronic Relations}
      \begin{itemize}
        \item \textbf{Agreement} iff\\
              (entity$_1$ = entity$_2$) and (in\_what$_1$ = in\_what$_2$) and
              (time\_span$_1$ = time\_span$_2$) and (value$_1$ = value$_2$)
        \item \textbf{Near Agreement} iff\\
              (entity$_1$ = entity$_2$) and (in\_what$_1$ = in\_what$_2$) and
              (time\_span$_1$ = time\_span$_2$) and (value$_1$ $\approx$ value$_2$)
        \item \textbf{Disagreement} iff\\
              (entity$_1$ = entity$_2$) and (in\_what$_1$ = in\_what$_2$) and
              (time\_span$_1$ = time\_span$_2$) and (value$_1$ $\neq$ value$_2$)
      \end{itemize}
  \end{quote}\normalsize
  \caption{Specifications of Relations}\label{fig:relSpecs}
\end{figure}

A question that can arise is the following: \textit{How does time affect the
relations you create?} To answer that question, imagine having two identical
messages, in different documents. If the documents have chronological distance
zero, then we have an \textsl{agreement} relation. If the messages come from
the same source but have chronological distance 1, then we have a
\textsl{stability} relation. Finally, if the messages come from different
sources and have chronological distance more than one, then we have no relation
at all. Thus, indeed, time does affect the relations.

\paragraph{An Example} At this point we would like to give a more concrete
example. Two sources, A and B, for a particular match, describe the performance
of a player as follows:
\begin{description}
  \item[A] \small The performance of Nalitzis, for the whole match was mediocre.\normalsize
  \item[B] \small In general, we can say that Nalitzis performed modestly,
           throughout the match.\normalsize
\end{description}
The messages that represent those two sentences are the following:
\begin{description}
  \item[A] \scriptsize\texttt{performance (Nalitzis, general, whole\_match, 50)} \normalsize
  \item[B] \scriptsize\texttt{performance (Nalitzis, general, whole\_match, 50)}\normalsize
\end{description}
The number 50 represents the mediocre performance of the player, since the
\texttt{degree} is realized as an integer in the scale of 0 to 100. According
to the specifications of the relations (see Figure~\ref{fig:relSpecs}) we would
have an \textsl{Agreement} synchronic relations between those two messages. In
the next game we have the following description:
\begin{description}
  \item[A] \small Nalitzis shown an excellent performance throughout the game. \normalsize
\end{description}
The message that results from this sentence is the following:
\begin{description}
  \item[A] \scriptsize\texttt{performance (Nalitzis, general, whole\_match, 100)} \normalsize
\end{description}
Now, between the two messages from source A we have a \textsl{Positive
Graduation} diachronic relation.

\subsection{The System}
\begin{figure}[thb]
  \begin{center}
    \ifpdf
      \includegraphics[scale=0.7,angle=270]{images/grid.pdf}
    \else
      \includegraphics[scale=0.7]{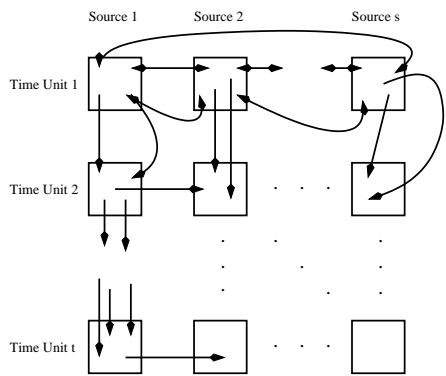}
    \fi
  \end{center}
  \caption{The Grid structure with Synchronic and Diachronic Relations}
  \label{fig:grid}
\end{figure}

Our summarization system is a query-based one, since the summary is an answer
to a natural language query that a user has posed. Such queries concern the
evolution of several events in the domain. In order to create the summaries we
have to extract, from the documents, the messages with their arguments, and the
relations that connect them, and subsequently organize them into a structure
which we call a \emph{grid} (see Figure~\ref{fig:grid}). This grid reflects
exactly the fact that the domain that we have used in this case study exhibits
linear evolution. If we take a horizontal ``slice'' of the grid, then we will
have descriptions of events from all the sources, for a particular time unit.
If, on the other hand, we take a vertical ``slice'' of the grid, then we have
the description of the evolution of an event from a particular source.

In order to extract the messages from the documents, our system employs an \IE
subcomponent. Relations between the messages are identified according to the
conditions associated with each one. After the user has issued the query, the
system identifies the various messages that are relevant to this query, as well
as the relations that connect them. Thus, in essence the system \emph{extracts
a subgrid} from the original grid which is, in fact, the answer to the user
query. This subgrid is passed to a \nlg subcomponent which creates the final
summary.

\subsubsection{Messages Extraction}
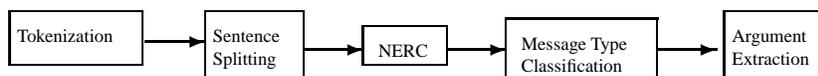
\begin{figure*}[thb]
  \begin{center}
    \setlength{\unitlength}{4144sp}%
\begingroup\makeatletter\ifx\SetFigFont\undefined%
\gdef\SetFigFont#1#2#3#4#5{%
  \reset@font\fontsize{#1}{#2pt}%
  \fontfamily{#3}\fontseries{#4}\fontshape{#5}%
  \selectfont}%
\fi\endgroup%
\begin{picture}(4929,429)(259,197)
\thinlines
{\color[rgb]{0,0,0}\put(271,299){\framebox(788,315){}}
}%
\thicklines
{\color[rgb]{0,0,0}\put(1081,434){\vector( 1, 0){360}}
}%
\thinlines
{\color[rgb]{0,0,0}\put(1441,209){\framebox(585,405){}}
}%
\thicklines
{\color[rgb]{0,0,0}\put(2026,389){\vector( 1, 0){360}}
}%
\thinlines
{\color[rgb]{0,0,0}\put(2386,299){\framebox(495,225){}}
}%
\thicklines
{\color[rgb]{0,0,0}\put(2881,389){\vector( 1, 0){360}}
}%
\thinlines
{\color[rgb]{0,0,0}\put(3241,209){\framebox(900,360){}}
}%
\thicklines
{\color[rgb]{0,0,0}\put(4141,389){\vector( 1, 0){360}}
}%
\thinlines
{\color[rgb]{0,0,0}\put(4546,209){\framebox(630,405){}}
}%
\put(1486,299){\makebox(0,0)[lb]{\smash{{\SetFigFont{7}{8.4}{\familydefault}{\mddefault}{\updefault}{\color[rgb]{0,0,0}Splitting}%
}}}}
\put(2476,344){\makebox(0,0)[lb]{\smash{{\SetFigFont{7}{8.4}{\familydefault}{\mddefault}{\updefault}{\color[rgb]{0,0,0}NERC}%
}}}}
\put(3331,389){\makebox(0,0)[lb]{\smash{{\SetFigFont{7}{8.4}{\familydefault}{\mddefault}{\updefault}{\color[rgb]{0,0,0}Message Type}%
}}}}
\put(3331,254){\makebox(0,0)[lb]{\smash{{\SetFigFont{7}{8.4}{\familydefault}{\mddefault}{\updefault}{\color[rgb]{0,0,0}Classification}%
}}}}
\put(4591,434){\makebox(0,0)[lb]{\smash{{\SetFigFont{7}{8.4}{\familydefault}{\mddefault}{\updefault}{\color[rgb]{0,0,0}Argument}%
}}}}
\put(4591,299){\makebox(0,0)[lb]{\smash{{\SetFigFont{7}{8.4}{\familydefault}{\mddefault}{\updefault}{\color[rgb]{0,0,0}Extraction}%
}}}}
\put(316,434){\makebox(0,0)[lb]{\smash{{\SetFigFont{7}{8.4}{\familydefault}{\mddefault}{\updefault}{\color[rgb]{0,0,0}Tokenization}%
}}}}
\put(1486,434){\makebox(0,0)[lb]{\smash{{\SetFigFont{7}{8.4}{\familydefault}{\mddefault}{\updefault}{\color[rgb]{0,0,0}Sentence}%
}}}}
\end{picture}%
  \end{center}
  \caption{The message extraction subsystem}\label{fig:msgSystem}
\end{figure*}

This subsystem was developed using the \textit{Ellogon} text engineering
platform.\footnote{\url{www.ellogon.org}} Its architecture is depicted in
Figure~\ref{fig:msgSystem}. It involves the following processing stages.

\paragraph{Preprocessing.}
This stage includes the tokenization, sentence splitting and the \nerc
sub-stages. During \nerc, we try to identify the \nes in the documents and
classify them into the categories that the ontology proposes.

\medskip\noindent The next two processing stages are the core of message
extraction. In the first one we try to identify the \emph{type} of each
extracted message, while in the second we try to fill its \emph{argument
values}.

\paragraph{Message Classification.}
Concerning the identification of the message types, we approached it as a
classification problem. From a study that we carried out, we concluded that in
most of the cases the mapping from sentences to messages was one-to-one, \ie in
most of the cases one sentence corresponded to one message. Of course, there
were cases in which one message was spanning more than one sentence, or that
one sentence was containing more than one message. We managed to deal with such
cases during the arguments' filling stage.

%

In order to perform our experiments we used a bag-of-words approach according
to which we represented each sentence as a vector from which the stop-words and
the words with low frequencies (four or less) were removed. The features used
are divided into two categories: \emph{lexical} and \emph{semantic}. As lexical
features we used the words of the sentences both stemmed and unstemmed. As
semantic features we used the \emph{NE types} that appear in the sentence. Of
course, in order to perform the training phase of the experiments, in each of
the vectors we appended the \emph{class} of the sentence, \ie the type of
message; in case a sentence did not corresponded to a message we labeled that
vector as belonging to the class \textsl{None}. This resulted into \emph{four}
series of vectors and corresponding experiments that we performed.

In order to perform the classification experiments we used the WEKA platform
\cite{WEKA.00}. The \ml algorithms that we used where three: \emph{Na\"{i}ve
Bayes, LogitBoost} and \emph{SMO}. For the last two algorithms, apart from the
default configuration, we performed some more experiments concerning several of
their arguments. Thus for the LogiBoost we experimented with the number of
iterations that the algorithm performs and for the SMO we experimented with the
complexity constant, with the exponent for the polynomial kernel and with the
gamma for the RBF kernel. For each of the above combinations we performed a
\emph{ten-fold cross-validation} with the annotated corpora that we had. The
results of the above experiments are presented in
Table~\ref{table:classificationResults}.

\begin{table*}[thb]
  \centering\scriptsize
  \begin{tabularx}{\textwidth}{|l|l|X||l|l|X|}
    \hline
    \multicolumn{2}{|c|}{\textbf{Classifier}} & \textbf{Correctly Classified Instances} &
    \multicolumn{2}{|c|}{\textbf{Classifier}} & \textbf{Correctly Classified Instances} \\
    \hline\hline
    \multicolumn{3}{|c||}{Without NE types} & \multicolumn{3}{|c|}{Including NE types} \\
    \hline
              & Na\"{i}ve Bayes             & 60.6693 \% &           & Na\"{i}ve Bayes             & 63.8286 \% \\
              & LogitBoost default          & 72.7443 \% &           & LogitBoost default          & 78.0991 \% \\
              & LogitBoost $I=5$            & 71.8876 \% &           & LogitBoost $I=5$            & 76.1981 \% \\
    stemmed   & LogitBoost $I=15$           & 72.2892 \% & stemmed   & LogitBoost $I=15$           & 78.2062 \% \\
              & SMO default                 & 73.6011 \% &           & SMO default                 & 75.9839 \% \\
              & SMO $C=0.5\ E=0.5\ G=0.001$ & 68.9692 \% &           & SMO $C=0.5\ E=0.5\ G=0.001$ & 72.5301 \% \\
              & SMO $C=1.5\ E=1.5\ G=0.1$   & 74.4578 \% &           & SMO $C=1.5\ E=1.5\ G=0.1$   & 75.7965 \% \\
    \hline
              & Na\"{i}ve Bayes             & 62.2758 \% &           & Na\"{i}ve Bayes             & 64.2035 \% \\
              & LogitBoost default          & 75.8768 \% &           & LogitBoost default          & 78.9023 \% \\
              & LogitBoost $I=5$            & 74.9398 \% &           & LogitBoost $I=5$            & 77.4565 \% \\
    unstemmed & LogitBoost $I=15$           & 76.6533 \% & unstemmed & LogitBoost $I=15$           & 79.4645 \% \\
              & SMO default                 & 79.2503 \% &           & SMO default                 & 79.6252 \% \\
              & SMO $C=0.5\ E=0.5\ G=0.001$ & 75.2343 \% &           & SMO $C=0.5\ E=0.5\ G=0.001$ & 76.8675 \% \\
              & SMO $C=1.5\ E=1.5\ G=0.1$   & 77.9920 \% &           & SMO $C=1.5\ E=1.5\ G=0.1$   & 78.5007 \% \\

    \hline
  \end{tabularx}
  \normalsize\caption{The results from the classification experiments}
  \label{table:classificationResults}
\end{table*}

Taking a look at that table there are several remarks that we can make.
Firstly, the LogitBoost and the SMO classifiers that we used outperformed, in
all the cases, the Na\"{i}ve Bayes which was our baseline classifier. Secondly,
the inclusion of the NE types in the vectors gave a considerable enhancement to
the performance of all the classifiers. This is rather logical, since almost
all the messages contain in their arguments \nes. The third remark, concerns
the stemmed and the unstemmed results. As we can see from the table, the
algorithms that used vectors which contained unstemmed words outperformed the
corresponding algorithms which used vectors whose words had been stemmed. This
is rather counterintuitive, since in most of the cases using stemming one has
better results.

Ultimately, the algorithm that gave the best results, in the experiments we
performed, was the SMO with the default configuration for the unstemmed vectors
which included information on the NE types. This classifier managed to
correctly classify 2974 out of 3735 messages (including the \textsl{None}
class) or about 80\% of the messages. Thus, we integrated this trained
classifier in the message extraction subsystem, which you can see in
Figure~\ref{fig:msgSystem}.

\paragraph{Arguments' Filling}
In order to perform this stage we employed several domain-specific heuristics.
Those heuristics take into account the constraints of the messages, if they do
have. As we noted above, one of the drawbacks of our classification approach is
that there are some cases in which we do not have a one-to-one mapping from
sentences to messages. During this stage of message extraction we used
heuristics to handle many of these cases.

In Table~\ref{table:gridStatistics} we show the final performance of the
subsystem as a whole, when compared against manually annotated messages on the
corpora used. Those measures concern only the message types. As you can see
from that table although the vast majority of the messages extracted are
correct, these represent 68\% of all the messages.

\begin{table}[thb]
  \centering
  \begin{tabular}{ll}
    Precision &: \quad 91.1178\\
    Recall    &: \quad 67.7810\\
    F-Measure &: \quad 77.7357\\
  \end{tabular}
  \caption{Evaluation of the messages' extraction stage}
  \label{table:gridStatistics}
\end{table}

\subsubsection{Extraction of Relations}
As is evident from Figure~\ref{fig:relSpecs}, once we have identified the
messages in each document and we have placed them in the appropriate position
in the grid, then it is fairly straightforward, through their specifications,
to identify the cross-document relations among the messages.

In order to achieve that, we implemented a system which was written in Java.
This system takes as input the extracted messages with their arguments from the
previous subsystem and it is responsible for the incorporation of the ontology,
the representation of the messages and the extraction of the synchronic and
diachronic cross-document relations. Ultimately, through this system we manage
to represent the \emph{grid}, which carries an essential role for our
summarization approach.

The reason for this is that since our approach is a query based one, we would
like to be able to pose queries and get the answers from the grid. The system
that we have created implements the API through which one can pose queries to
the grid, as well as the mechanism that extracts from the whole grid structure
the appropriate messages and the relations that accompany them, which form an
answer to the question. Those extracted messages and relations form a sub-grid
which can then be passed to an \nlg system for the final creation of the
summary.

Concerning the statistics of the extracted relation, these are presented in
Table~\ref{table:relStatistics}. The fact that we have lower statistical
measures on the relations, in comparison with the message types, can be
attributed to the argument extraction subsystem, which does not perform as well
as the message classification subsystem.

\begin{table}[thb]
  \centering
  \begin{tabular}{ll}
    Precision &: \quad 89.0521\\
    Recall    &: \quad 39.1789\\
    F-Measure &: \quad 54.4168\\
  \end{tabular}
  \caption{Recall, Precision and F-Measure on the relations}
  \label{table:relStatistics}
\end{table}

As of writing this paper, everything has been implemented except the mechanism
that transforms the natural language queries to the API that will extract the
sub-grid. Additionally, we do not have a connection with an \nlg system, but
instead we have implemented some simple template-based mechanism.

\section{Related Work}\label{sec:related}
The work that we present in this paper is concerned with multi-document
summarization of events that evolve through time. Of course, we are not the
first to incorporate directly, or indirectly, the notion of time in our
approaches to summarization. \citeN{Lehnert.81}, for example, attempts to
provide a theory for what she calls \emph{narrative summarization}. Her
approach is based on the notion of ``plot units'', which connect \emph{mental
states} with several relations, and are combined into very complex patterns.
This approach is a single-document one and was not implemented. Recently,
\citeN{Mani.04} attempts to revive this theory of narrative summarization,
although he also does not provide any concrete computational approach for its
implementation.

From a different viewpoint, \citeN{Allan&al01} attempt what they call
\emph{temporal summarization}. In order to achieve that, they take the results
from a \tdt system for an event, and they put all the sentences one after the
other in a chronological order, regardless of the document that it belonged,
creating a stream of sentences. Then they apply two statistical measures
\emph{usefulness} and \emph{novelty} to each ordered sentence. The aim is to
extract those sentences which have a score over a certain threshold. This
approach does not take into account the document sources, and it is not
concerned with the evolution of the events; instead they try to capture novel
information.

As we have said, our work requires some domain knowledge which is expressed
through the ontology, and the messages' and relations' specifications. A system
which is based also on domain knowledge is \textsc{summons}
\cite{Radev&McKeown98,Radev99:PhD}. The domain knowledge for this system comes
from the specifications of the \muc conferences. This system takes as input
several \muc templates and, applying a series of operators, it tries to create
a baseline summary, which is then enhanced by various named entity descriptions
collected from the Internet. One can argue that the operators that
\textsc{summons} uses resemble our cross-document relations. This is a
superficial resemblance, since our relations are divided into \emph{synchronic}
and \emph{diachronic}, thus reporting similarities and differences in two
different directions: source and time. Additionally our system is a query-based
one.

Concerning now the use of relations, \cite{Salton&al97} for example, try to
extract paragraphs from a single document by representing them as vectors and
assigning a relation between the vectors if their similarity exceeds a certain
threshold. Then, they present various heuristics for the extraction of the best
paragraphs.

Finally, \citeN{Radev00} proposed the \cstlong which incorporated a set of 24
domain independent relations that exist between various textual units across
documents. In a later paper \citeN{Zhang&al02} reduce that set into 17
relations and perform some experiments with human judges. Those experiments
reveal several interesting results. For example, human judges annotate only
sentences, ignoring the other textual units (phrases, paragraphs, documents)
that the theory suggests. Additionally, we see a rather small inter-judge
agreement concerning the type of relation that connects two sentences.
\citeN{Zhang&al03} and \citeN{Zhang&Radev.04a} continue the work with some
experiments, during which they use \ml techniques to identify the
cross-document relations. We have to note here that although a general
\emph{pool} of cross-document relations might exist, we believe, in contrast
with \citeN{Radev00}, that those relations are dependent on the domain, in the
sense that one can choose from this pool the appropriate subset of relations
for the domain under consideration, possibly enhancing this subset with
completely domain specific relations that will suit ones needs. Another
significant difference from our work, is that our main goal is to create
summaries that show the evolution of an event, as well as the similarities or
differences that the sources have during the evolution of an event.

\section{Conclusions and Future Work}\label{sec:future}
The aim of this paper was to present our approach to the problem of
\emph{multi-document summarization of evolving events}. We divide the evolution
of the events into linear and non-linear. In order to tackle the problem, we
introduced cross-document relations which represent the evolution of the events
in two axes: \emph{synchronic} and \emph{diachronic}. Those relations connect
messages, which represent the main events of the domain, and are dependent on
the domain ontology. We also presented, through a case study, an implementation
for a linearly evolving domain, namely that of the descriptions of football
matches. The system we have built automatically extracts the messages and the
synchronic and diachronic relations from the text. A particular point of
concern is the recall (approximately 40\%) of the relations' extraction
sub-system, which is due to the heuristics used for the filling the arguments
of the messages. Apart from enhancing our heuristics, we also plan to study
their effect on the quality of the generated summary.

Currently we are working on a more complicated domain, namely that of events
with hostages, whose evolution, according to the specification that we gave in
the introduction of this paper, can be characterized as non-linear. The main
challenges in non-linear evolution concern the synchronic relations. A related
problem, which we investigate, is that of the \emph{temporal expressions} which
may make several messages refer back in time, in relation to the publication
time of the article that contains the messages. We also examine in depth the
role that time has on the relations. Additionally, we examine the existence of
relations between different message types. Concerning now the classification
experiments and the argument extraction, we intend to enhance them by adding
more semantic features incorporating also the Greek
WordNet.\footnote{\url{www.ceid.upatras.gr/Balkanet/resources.htm}}


\bibliographystyle{ranlp}
\begin{scriptsize}
\bibliography{ranlp05}
\end{scriptsize}

\end{document}